# MGSC: A MULTI-GRANULARITY CONSISTENCY FRAMEWORK FOR ROBUST END-TO-END ASR

Xuwen Yang[1]

[1]China University of Mining and Technology, Jiangsu, China
TS24170053A31 @cumt.edu.cn

**Abstract.** End-to-end ASR models, despite their success on benchmarks, often produce catastrophic semantic errors in noisy environments. We attribute this fragility to the prevailing 'direct mapping' objective, which solely penalizes final output errors while leaving the model's internal computational process unconstrained. To address this, we introduce the Multi-Granularity Soft Consistency (MGSC) framework, a model-agnostic, plug-and-play module that enforces internal self-consistency by simultaneously regularizing macro-level sentence semantics and micro-level token alignment. Crucially, our work is the first to uncover a powerful synergy between these two consistency granularities: their joint optimization yields robustness gains that significantly surpass the sum of their individual contributions. On a public dataset, MGSC reduces the average Character Error Rate by a relative 8.7% across diverse noise conditions, primarily by preventing severe meaning-altering mistakes. Our work demonstrates that enforcing internal consistency is a crucial step towards building more robust and trustworthy AI.

**Keywords:** Speech Recognition, Internal Consistency, Attention Alignment

## 1 INTRODUCTION

End-to-end ASR models, exemplified by the Conformer architecture, have demonstrated exceptional performance on benchmark datasets. This success, however, belies a critical vulnerability in noisy environments where these models are prone to semantically catastrophic errors, such as misinterpreting "disapprove" as "approve"[1]. Such failures can lead to severe consequences in safety-critical applications. We attribute this fragility to the prevailing Direct Mapping Paradigm, which treats the model as an open-loop system. In this paradigm, learning is driven exclusively by the discrepancy between the model's final output and the ground-truth transcript, leaving the internal computational process entirely unconstrained.

This lack of internal supervision fosters representational inconsistencies when the model confronts acoustic uncertainty. These inconsistencies manifest at two distinct granularities. At the macro-level, we identify semantic drift, a divergence between the encoder's global acoustic understanding and the decoder's generative intent. At the micro-level, we observe alignment chaos, where the attention mechanism fails to preserve the monotonic temporal structure of speech, resulting in corrupted decoding. Together, these internal conflicts systematically undermine the model's reliability.

Xuwen Yang

To address this fundamental limitation, we introduce the Multi-Granularity Soft Consistency (MGSC) framework, a novel, model-agnostic module designed to enforce internal self-consistency. Instead of replacing the existing paradigm, MGSC enhances it by integrating two concurrent regularization terms into the learning objective: (i) a sentence-level semantic constraint that aligns the encoder's global representation with that of the decoder, and ( ii) a token-level alignment constraint that promotes structural monotonicity.

The main contributions of this paper are threefold: 1) We identify and systematically analyze two distinct granularities of internal inconsistency—macro-level semantic drift and micro-level alignment chaos—as a root cause of catastrophic errors in robust ASR. 2) We design MGSC, a plug-and-play framework that, for the first time, unifies macro-semantic and micro-structural consistency constraints within a single, coherent learning objective. 3) We provide strong empirical evidence for a powerful synergistic effect between these two consistency constraints, demonstrating that their joint optimization yields robustness gains that significantly surpass the sum of their individual contributions.

## 2 RELATED WORK

Traditional methods for enhancing the robustness of ASR systems include data augmentation and robust training strategies. For instance, SpecAugment[2] improves model adaptation to noise by randomly masking the audio spectrogram, while large-scale weak supervision [3] significantly boosts model generalization across diverse scenarios by leveraging massive unlabeled datasets. However, recent research has increasingly focused on the internal structural integrity of models, particularly the alignment logic and semantic coherence between the encoder and decoder, to combat semantic drift and alignment chaos in noisy environments.

Speech-to-text conversion must adhere to temporal monotonicity. Hard-constraint methods, such as early monotonic attention mechanisms[4], enforce monotonicity by restricting the attention window, but at the cost of model flexibility and representation capacity. In contrast, soft-guidance methods guide monotonicity through regularization or attention matrix reconstruction. For example, supervised attention mechanisms [5] improve the correspondence between attention weights and true alignments by introducing a supervised attention loss, significantly enhancing the performance of sequence-to-sequence models. Furthermore, CR-CTC [6] strengthens the model's ability to capture temporal logic by introducing a consistency regularization term in the CTC model. Our $L_{align}$ regularization term, based on an expected alignment path and Hinge Loss, precisely penalizes logical regressions while tolerating reasonable pauses, thereby balancing computational efficiency and alignment accuracy, outperforming more complex multi-step reconstruction methods.

Regarding macro-level semantic consistency, recent studies have explored enhancing ASR's semantic understanding through semantic lattice rescoring and contextual biasing. For example, Transformer-based semantic lattice rescoring [7] significantly reduces the error rate of context-dependent words by re-evaluating word sequence probabilities. Similarly, Factual Consistency Oriented Speech Recognition [8] mitigates hallucination phenomena in noisy settings by optimizing the factual consistency score between model hypotheses and the ground truth. Additionally, contextual biasing methods [9] improve the recognition of rare words and named entities through early context injection and text perturbation. More recently, Lin [10] proposed a phrase-level contextualized ASR method that enhances semantic consistency through dynamic vocabulary prediction and activation. This approach significantly reduces the error rate on contextual phrases by optimizing phrase-level predictions, sharing a similar goal with our $L_{sentence}$ in pursuing global semantic coherence. However, the aforementioned methods often rely on specific contexts or external language models, making it difficult to systematically address noise-induced semantic drift. In contrast, our proposed MGSC framework constructs a closed-loop, self-reflective system through efficient micro-level alignment ( $L_{align}$ ) and an innovative macro-level semantic consistency constraint ($L_{sentence}$). By aligning global representations, $L_{sentence}$ imposes a high-level semantic consistency constraint between the model's acoustic encoder and text decoder. This constraint compels the model to learn a unified semantic representation that is robust to acoustic interference, such as noise, thereby effectively mitigating semantic drift and ensuring the final output remains consistent with the overall intent of the acoustic input.

Survey literature [11] indicates that attention mechanisms and consistency constraints[12] are central to current ASR robustness research. Our MGSC framework not only integrates these advancements but also provides a more comprehensive robustness solution through the synergistic interplay of multi-granularity consistencies, marking a paradigm shift from solely pursuing representation power to focusing on the model's internal cognitive self-consistency[13].

## 3    THE MGSC FRAMEWORK

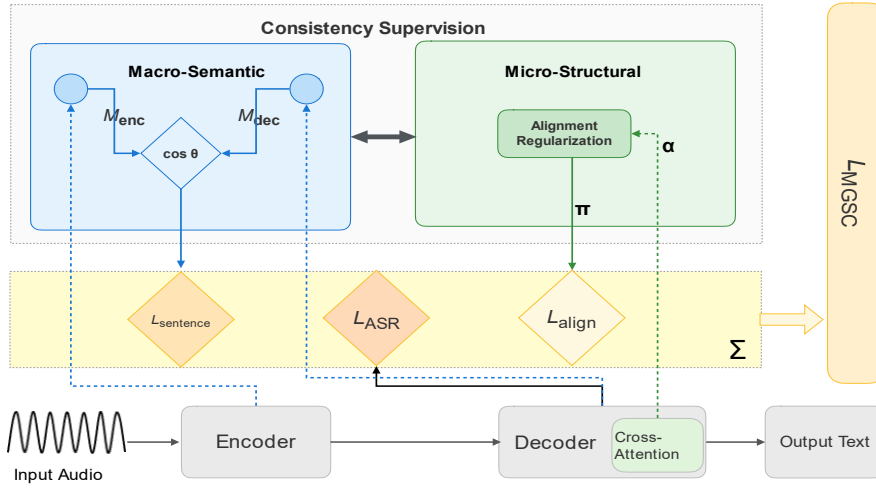

**Fig. 1.** Overall architecture of the MGSC framework.

We propose the Consistency-Driven Learning paradigm, designed to overcome the limitations of the traditional Direct Mapping approach by incorporating internal self-consistency into the learning objective. To this end, we designed the Multi-Granularity Soft Consistency (MGSC) framework, which imposes constraints on the model at both the macro-semantic and micro-structural levels via two parallel regularization terms.

The overall architecture of this framework is illustrated in Figure 1. It is built upon a standard encoder-decoder backbone, depicted by the grey components, on top of which we construct its core module: the Consistency Supervision Layer. This layer imposes internal consistency constraints in parallel at two distinct granularities: (1) at the macro-semantic level, shown in blue, we aim to ensure that the encoder's global understanding of the acoustic information remains semantically consistent with the decoder's generative intent. To achieve this, we extract global representations, $M_{enc}$ and $M_{dec}$, from the encoder and decoder, respectively, via the global average pooling operation represented by the blue circles. We then compute the consistency loss $L_{sentence}$ between these two representations. (2) Second, at the micro-structural level, shown in green, we impose a soft monotonicity constraint on the cross-attention mechanism to rectify temporal alignment errors that may occur during the decoding process. This yields the alignment loss $L_{align}$. Finally, these two consistency regularization losses, originating from different granularities, are combined with the standard ASR task loss $L_{ASR}$ through a weighted summation to form the final learning objective $L_{MGSC}$, depicted in yellow. In the following subsections, we will first present the overall learning objective and then detail the implementation of the macro-semantic and micro-structural consistency modules.

### 3.1  The Consistency-Driven Learning Objective

In standard end-to-end ASR training, the objective function $L_{ASR}$ is typically a weighted sum of the Cross-Entropy and Connectionist Temporal Classification (CTC)[14] losses:

$$L_{DirectMap} = L_{ASR}(f(X), Y) \tag{1}$$

where $f$ denotes the ASR model, and $X$ and $Y$ are the input acoustic features and target text, respectively. This objective solely drives the model to match external labels, neglecting the logical coherence of its internal reasoning process.

Our consistency-driven paradigm augments this objective with an internal consistency regularization term $R_{consistency}$. The MGSC framework decomposes this term into two independent components: a sentence-level semantic consistency loss $L_{sentence}$ and a token-level alignment consistency loss $L_{align}$. The final training objective is defined as:

$$L_{MGSC} = L_{ASR} + \lambda_{sent} \cdot L_{sentence} + \lambda_{align} \cdot L_{align} \tag{2}$$

Where $\lambda_{sent}$ $\lambda_{align}$ and are hyperparameters that balance the respective loss terms. This joint optimization objective guides the model to not only learn the correct transcription but also to construct an internally logical and self-consistent cognitive system.

### 3.2  Sentence-Level Semantic Consistency

To mitigate the "macro-level semantic drift" that occurs under noisy conditions—a divergence between the encoder's global acoustic understanding and the decoder's generated text—we introduce a sentence-level semantic consistency constraint. Its core hypothesis is that for a robust model, the encoder and decoder, despite their different functions, should maintain consistent global semantic representations for the same utterance in the latent space[15].

To achieve this, we first extract the model's global acoustic and semantic representations. The acoustic-side global representation, $M_{enc}$, is obtained by applying average pooling over the time dimension of the encoder's final layer output $H_{enc} \in R^{T_{in} \times D}$.

$$M_{enc} = \frac{1}{T_{in}} \sum_{t=1}^{T_{in}} h_{enc,t} \qquad h_{enc,t} \in R^D \tag{3}$$

Where $T_{in}$ is the input sequence length and $D$ is the hidden dimension. Similarly, the text-side global representation, $M_{dec}$, is derived by averaging all hidden states $H_{dec,i}$ generated by the decoder during sentence production:

$$M_{dec} = \frac{1}{T_{out}} \sum_{t=1}^{T_{out}} h_{dec,i} \qquad h_{dec,i} \in R^D \tag{4}$$

Since acoustic variations like noise primarily affect the magnitude of representation vectors, while core semantic information is more robustly encoded in their direction, metrics sensitive to magnitude, such as the L2 distance, are suboptimal. We therefore employ cosine similarity to measure the consistency between these two global representations[16]. The sentence-level semantic consistency loss, $L_{sentence}$, is defined as their negative cosine similarity :

$$L_{sentence} = 1 - cos(M_{enc}, M_{dec}) = 1 - \frac{M_{enc} \cdot M_{dec}}{\|M_{enc}\|\|M_{dec}\|} \quad (5)$$

By minimizing this loss, we compel the model to learn a global semantic representation that is invariant to acoustic perturbations. This representational consistency imposes a powerful global semantic constraint, which fundamentally suppresses semantic drift by continuously calibrating the model's overall generative intent.

### 3.3 Token-Level Alignment Consistency

To correct micro-level alignment chaos, we introduce a regularization term that promotes temporal monotonicity in the attention mechanism. Unlike rigid constraints that impair model flexibility, our soft guidance approach selectively penalizes illogical attention regressions (look-backs) while tolerating necessary pauses, such as those for long vowels or inter-word silences[17].

We first quantify the alignment position at each decoding step. For the Transformer's cross-attention weight matrix $\alpha \in R^{T_{out} \times T_{in}}$, we compute an expected alignment path [18]vector $\pi \in R^{T_{out}}$, where each element $\pi_i$ represents the weighted average index of the input frame attended to by the $i$-th output token:

$$\pi_i = \sum_{j=1}^{T_{in}} \alpha_{i,j} \cdot (j-1) \quad (6)$$

This vector $\pi$ delineates the alignment trajectory throughout the decoding process. The ideal monotonic non-decreasing property requires $\pi_i < \pi_{i-1}$ for all $i > 1$.

To softly enforce this property, we employ the Hinge Loss[19]to construct the token-level alignment consistency loss $L_{align}$. This loss incurs a penalty only when an alignment regression occurs (i.e., $\pi_i < \pi_{i-1}$ ):

$$L_{align} = \frac{1}{T_{out} - 1} \sum_{i=2}^{T_{out}} max(0, \pi_{i-1} - \pi_i) \quad (7)$$

By design, this formulation is discriminative, applying a penalty only to non-monotonic alignments ($\pi_i < \pi_{i-1}$) in proportion to the magnitude of the regression. Consequently, this temporal regularization selectively penalizes instances where the attention regresses, while remaining permissive of valid pauses or forward alignments. This design enables the effective correction of structural alignment errors while preserving the flexibility of the attention mechanism and the model's representational capacity[20].

## 4 EXPERIMENTS

### 4.1 The Consistency-Driven Learning Objective

We performed experiments on the public AISHELL-1 [21]dataset. To evaluate robustness, we created noisy test sets by mixing the official test data with the NOISE-92 [22]corpus at five SNRs: 0, 2.5, 5, 7.5, and 10 db. CER is our primary evaluation metric.

Our baseline is a standard Conformer implemented via the WeNet toolkit[23] It is trained with a hybrid CTC/Attention[24] loss, making it a representative example of the "direct mapping" paradigm.

To balance the loss terms, our MGSC framework employs a Homoscedastic uncertainty-based loss balancing mechanism[25] instead of fixed hyperparameters, which treats the weights $\lambda_{sentence}$ and $\lambda_{align}$ as learnable parameters.

### 4.2 Quantitative Result

The efficacy of the MGSC framework is validated through a comprehensive ablation study, with results detailed in Table 1. The quantitative data unequivocally demonstrates the superiority of our consistency-driven approach over the baseline.

Table 1. Ablation study (CER %).

| System | Clean | 0db | 2.5db | 5db | 7.5db | 10db | Noisy avg |
|---|---|---|---|---|---|---|---|
| Baseline | 4.76 | 21.52 | 14.47 | 10.03 | 7.37 | 6.76 | 12.08 |
| $+L_{align}$ | 4.78 | 21.47 | 14.28 | 9.83 | 7.42 | 6.73 | 11.95 |
| $+L_{sentence}$ | 4.71 | 19.32 | 14.17 | 9.62 | 7.25 | 6.72 | 11.42 |
| +MGSC | 4.64 | 18.87 | 13.39 | 9.36 | 7.02 | 6.52 | 11.03 |

The core finding of this study lies in the unified action of the two constraints. The full MGSC framework achieved the lowest Character Error Rate (CER) under all test conditions, attaining 4.64% on the clean dataset and an average of 11.03% in noisy environments. This result provides strong evidence that while macro-semantic consistency is crucial, its full potential can only be unlocked when grounded in structurally robust and coherent micro-level alignments. This performance gain, which surpasses the sum of the contributions from the individual components, clearly reveals a powerful synergistic effect between the multi-granularity consistencies.

To further analyze the contribution of each constraint, we visualize the relative CER reduction in Figure 2.

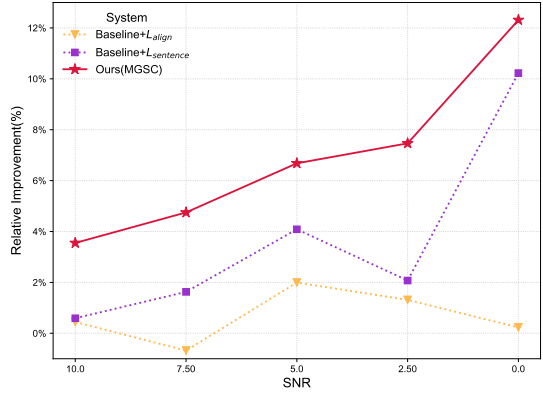

**Fig. 2.** Relative CER Reduction vs Baseline

This figure clearly demonstrates that the performance advantage of our full MGSC model becomes more pronounced as the noise level increases. This performance gain significantly surpasses the simple summation of the effects from the individual constraints. This phenomenon provides compelling quantitative evidence for the powerful synergistic effect we propose between micro-structural alignment and macro-semantic consistency.

### 4.3 Quantitative Analysis

**Attention Mechanism Analysis: Reconstruction of Micro-structural Consistency.** To provide qualitative insights into the framework's mechanism, we visualized the cross-attention maps of the baseline and MGSC models (Figure 3). The baseline model (Figure 3a) exhibits pronounced alignment chaos, characterized by a diffuse attention distribution and logically disruptive regressions. This confusion in the internal reasoning process directly contributes to decoding failures in noisy environments.

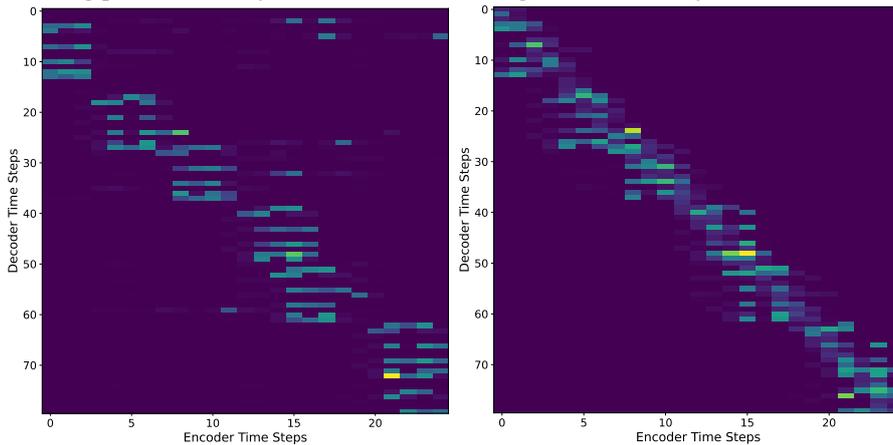

**Fig. 3.** A comparison of cross-attention maps between the baseline model and our proposed MGSC model

In stark contrast, the MGSC model (Figure 3b), guided by our token-level consistency constraint, produces a sharply focused and strictly monotonic alignment path. This constraint functions as a temporal ruler, remedying structural flaws in the attention mechanism. This restored micro-level integrity provides a stable foundation upon which macro-level semantic consistency can be reliably established, visually corroborating the synergistic effect observed in our quantitative results and underscoring the importance of internal self-consistency for model robustness.

**Latent Space Representation Analysis: Enforcing Macro-Semantic Consistency.**
To further investigate the impact of the $L_{sentence}$ constraint on the model's internal representations at the macro level, we visualized the global acoustic representations from the encoder $M_{enc}$ and the global semantic representations generated by the decoder $M_{dec}$ using the t-SNE[26] dimensionality reduction technique.

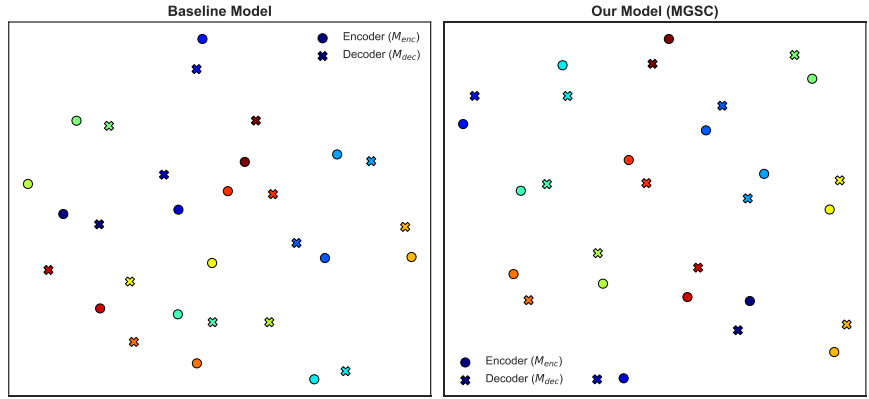

**Fig. 4.** Paired visualization of global representations $M_{enc}$ and $M_{dec}$ in the t-SNE space.

Figure 4 visually contrasts the impact of the $L_{sentence}$ constraint on the model's internal representations. In the baseline model, shown in Figure 4a, the acoustic representation $M_{enc}$ and the semantic representation $M_{dec}$ for the same input exhibit significant spatial separation in the latent space, revealing a severe semantic inconsistency between the encoder and decoder. In sharp contrast, the $L_{sentence}$ constraint in our MGSC model, depicted in Figure 4b, effectively pulls these representation pairs together to form several tightly co-located clusters. This result provides compelling evidence that our method successfully forces the model to learn a shared and noise-robust semantic space. It is this enforced macro-level consistency that fundamentally mitigates semantic drift and enhances the model's reliability.

**Error Attribution Analysis.**

Xuwen Yang

To further assess the practical impact of MGSC on recognition quality, we performed a manual attribution analysis on 500 error samples generated by the baseline model on the 0dB test set. Errors were classified into two categories: High-Severity Semantic Errors, which distort the core meaning, and Low-Severity Lexical Errors, which do not.

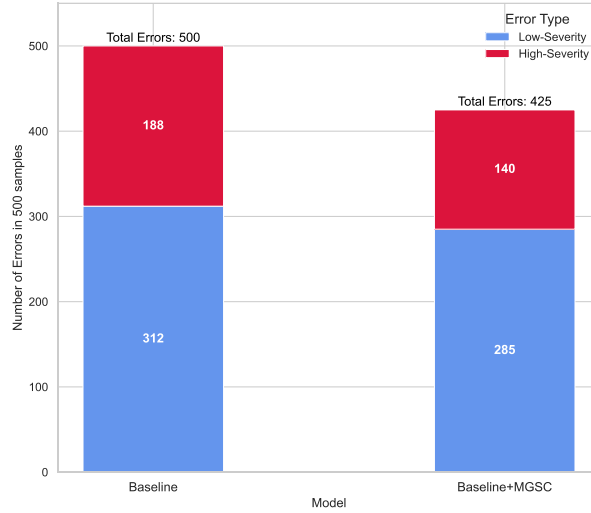

**Fig. 5.** Comparison of Error Types

The results are presented in Figure 5. While MGSC reduced the total number of errors from 500 to 425, its more critical contribution lies in drastically reducing the count of High-Severity Semantic Errors from 188 to 140—a relative reduction of 25.5%. This analysis provides compelling evidence that the primary advantage of our proposed consistency constraints is the significant mitigation of catastrophic semantic errors, effectively shifting the model's failure mode towards more benign lexical mistakes.

## 5     Conclusion

In this work, we addressed the fragility of end-to-end ASR models by targeting their internal computational inconsistencies. We introduced the MGSC framework, a model-agnostic module that enforces self-consistency at two distinct granularities: macro-level semantics and micro-level structural alignment. Our central finding is the discovery of a powerful synergy between these two constraints; their joint optimization yields robustness gains that substantially exceed their individual effects. By shifting the focus from an exclusive reliance on input-output mapping towards ensuring internal integrity, our work represents a concrete step towards more reliable and trustworthy AI systems. We believe the underlying principle of enforcing multi-granularity consistency holds promise for other sequence-to-sequence tasks and offers a new perspective on building more explainable models.